\documentclass[runningheads]{llncs}

\usepackage{eccv}

\usepackage{eccvabbrv}

\usepackage{graphicx}
\usepackage{booktabs}
\usepackage{multicol}
\usepackage{multirow}
\usepackage[accsupp]{axessibility}  

\usepackage{hyperref}

\usepackage{orcidlink}
\begin{document}

\title{MASS: Motion-Aligned Selective Scan for Refinement in Flow-Based Video Frame Interpolation} 

\titlerunning{MASS: Motion-Aligned Selective Scan for Refinement in VFI}

\author{Jun-Sang Yoo\orcidlink{0000-0001-9053-699X} \and
Seung-Won Jung\orcidlink{0000-0002-0319-4467}\thanks{Corresponding author.}}

\authorrunning{J.-S Yoo and S.-W. Jung}

\institute{Department of Electrical Engineering, Korea University, Seoul, Korea 
\email{\{junsang7777,swjung83\}@korea.ac.kr}}

\maketitle

\begin{abstract}
Video frame interpolation (VFI) remains a challenging task, particularly when dealing with large, non-linear motions and complex occlusions. While flow-based methods are prevalent, they often struggle with ambiguous correspondences. Recent VFI methods based on selective State Space Models (SSMs) are still limited by static grid-based scanning that misaligns with physical motion. In this paper, we propose Motion-Aligned Selective Scan (MASS), a novel framework that reformulates feature scanning from static spatial grids to dynamic motion trajectories. 
MASS builds a feature sequence along each pixel’s flow-guided trajectory and aggregates it with an SSM.
Specifically, we introduce a learnable non-linear path integration to approximate complex curved trajectories via residual velocity updates, and a velocity-aware SSM that dynamically adjusts the sampling budget and step size based on motion magnitude. 
This adaptive strategy allocates denser sampling to fast-motion regions while keeping static regions efficient.
Furthermore, the aggregated states guide a refinement module to rectify intermediate flows and masks in an end-to-end manner. Extensive experiments indicate that MASS achieves highly competitive overall performance on standard benchmarks, establishing state-of-the-art results particularly in challenging scenarios with large displacements and complex dynamics.
\end{abstract}

\section{Introduction}\label{sec:intro}

Video frame interpolation (VFI) aims to synthesize an intermediate frame $I_t$ at time $t\in(0,1)$ from two consecutive frames $I_0$ and $I_1$, enabling applications such as slow-motion generation~\cite{jiang2018super, xiang2020zooming, reda2022film}, video compression~\cite{wu2018vcii,tan2025hint}, and novel view synthesis~\cite{Deepstereo,LearningViewSynthesis,liu2024video}. 
While recent works~\cite{danier2024ldmvfi,yoo2023video,zhang2025eden} also explore diffusion or higher-level object/foreground cues to improve the perceptual fidelity, flow-based approaches~\cite{kong2022ifrnet,huang2022real,bao2019depth,lee2020adacof,park2021asymmetric} remain a prevalent paradigm for enforcing temporal alignment by explicitly modeling inter-frame motion.
They estimate intermediate flows toward time $t$, warp the input frames, and reconstruct $I_t$ by blending the warped frames with a synthesis network.

Despite their success, estimating intermediate flow is inherently challenging because the intermediate frame is unobserved. Even with multi-scale or iterative estimators~\cite{teed2020raft,sun2018pwc}, large displacement, occlusion/disocclusion, thin structures, and repetitive textures often lead to ambiguous correspondences, which degrade warping-based reconstruction. 
To address this, recent methods~\cite{liu2024sparse,reda2022film} such as SGM-VFI explicitly strengthen the correspondence stage by introducing global matching cues. By establishing sparse global correspondences between $I_0$ and $I_1$ and converting them into intermediate-flow compensation, these methods improve robustness under large motion.
However, correspondence-centric approaches remain insufficient in regions where reliable matches are fundamentally unavailable, such as disocclusions and motion boundaries, forcing the interpolation pipeline to operate with only approximate flow estimates.

In parallel, another line of work explores global modeling architectures, including Transformers~\cite{vaswani2017attention} and selective state space models (SSMs)~\cite{gu2024mamba} for VFI~\cite{zhang2024vfimamba, jeong2025lc, lu2022video, zhang2023extracting, li2023amt}. These models offer an efficient mechanism for long-range context aggregation. However, such designs still depend on the reliability of the underlying motion hypothesis and typically constrain aggregation to spatially defined windows for efficiency, which can be challenged by occlusion and highly non-linear motion.
Furthermore, most existing SSM-based VFI methods~\cite{zhang2024vfimamba, jeong2025lc} serialize features using scan orders defined on static grids (e.g., raster or local windows). 
For VFI, the most informative context for a target pixel often lies along its physical motion path rather than in a static spatial neighborhood. This mismatch motivates us to replace grid-defined serialization with flow-guided trajectory serialization, better matching the continuous, motion-driven nature of the task.

\begin{figure}[t]
\centering
\includegraphics[width=\linewidth]{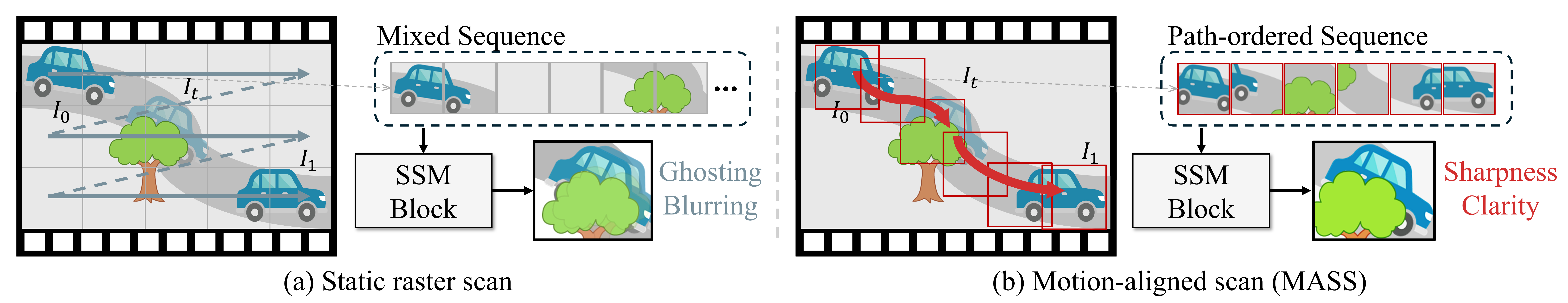}
\caption{Comparison of feature serialization strategies for VFI. (a) Static raster scanning flattens unaligned grids into mixed sequences, causing ghosting artifacts. (b) MASS extracts features along surrogate trajectories forming motion-consistent sequences and reducing misalignment in dynamic regions.}
\label{fig_first}
\end{figure}

In this paper, we introduce \textbf{Motion-Aligned Selective Scan (MASS)}, which shifts the scan order from static spatial grids to dynamic motion trajectories, turning per-pixel interpolation into trajectory-wise sequence modeling as illustrated in Fig.~\ref{fig_first}. Given coarse intermediate flows, MASS constructs a flow-guided surrogate trajectory for each pixel. Features sampled along this trajectory are flattened into a one-dimensional sequence. By aligning serialization with the motion hypothesis, MASS aggregates context along flow-guided trajectories rather than a fixed spatial grid, reducing interference from motion-inconsistent neighbors.

Specifically, MASS incorporates two mechanisms to overcome the limitations of linear interpolation: learnable velocity and adaptive sequence length. Real-world motion is rarely perfectly linear. To address this, MASS updates the trajectory state at each step using a learnable residual, enabling the model to approximate complex, curved paths. 
We vary the number of sampled points with motion magnitude, so longer trajectories are sampled more densely.
To prevent temporal aliasing and preserve transient details in fast-moving regions, our model assigns longer sequence lengths (i.e., denser sampling along the trajectory) to pixels with large displacements, while static areas are processed with shorter sequences for efficiency.

To demonstrate the effectiveness of our framework, we evaluate its performance on various standard benchmarks. The results show that MASS achieves state-of-the-art performance, particularly exhibiting superior robustness in complex scenarios with large displacements and occlusions without incurring significant computational overhead.

Our contributions are summarized as follows:
\begin{itemize}
    \item We propose Motion-Aligned Selective Scan (MASS), which replaces static grid serialization with flow-guided, learnable non-linear trajectories to better capture the continuous dynamics of frame interpolation.
    \item We introduce a motion-adaptive discretization strategy that dynamically adjusts sequence sampling and SSM step sizes, alongside a refinement module that uses forward-backward context discrepancies to correct intermediate motions.
    \item Extensive experiments show MASS is highly competitive across standard benchmarks, achieving the most significant gains in challenging scenarios with large displacements and high-resolution inputs.
    
\end{itemize}

\section{Related Work}\label{sec:rel}

\subsection{Flow-based Video Frame Interpolation}
Flow-based VFI synthesizes an intermediate frame by estimating intermediate-time correspondences and visibility cues (e.g., blending/occlusion masks), and reconstructing $I_t$ via warping and fusion~\cite{jiang2018super, bao2019depth}.
A major line of work derives these flows from bidirectional estimations~\cite{niklaus2020softmax,liu2020enhanced}, while another line directly predicts them via dedicated supervision~\cite{huang2022real,kong2022ifrnet,jin2023unified}.
Representative models like RIFE~\cite{huang2022real} establish highly efficient warp-and-merge paradigms, whereas IFRNet~\cite{kong2022ifrnet} emphasizes the intrinsic difficulty of intermediate flow estimation due to the absence of the target frame.

To strengthen inter-frame interaction and capture broader context, attention-augmented methods~\cite{lu2022video, zhang2023extracting} have been widely adopted.
VFIFormer~\cite{lu2022video} and EMA-VFI~\cite{zhang2023extracting} demonstrate that cross-frame attention effectively balances detail preservation and global reasoning.
TTVFI~\cite{TTVFI} introduces endpoint-conditioned linear motion modeling in Transformers to mitigate warping inconsistencies; however, it models trajectories simply as linear displacements between two endpoints and constrains its attention mechanism to static spatial windows.

Nevertheless, large displacement and high-resolution scenarios remain challenging, as correspondence errors can be amplified by warping and become difficult to correct.
Several works address this through scalable training strategies (e.g., XVFI~\cite{sim2021xvfi}), scale-robust estimation (e.g., FILM~\cite{reda2022film}), or stronger correlation-and-refinement schemes (e.g., AMT~\cite{li2023amt}).
Other approaches expand the effective matching range by incorporating bilateral correlation modeling~\cite{park2021asymmetric, park2023biformer} or sparse global matching compensation~\cite{liu2024sparse}.

Rather than introducing additional matching branches or relying on fixed interaction patterns, we revisit refinement from the perspective of motion-consistent aggregation. Our goal is to construct representations that remain informative even under local correspondence uncertainties by explicitly aggregating features along non-linear motion trajectories.

\subsection{State Space Models for Vision and VFI}
SSMs~\cite{gu2021s4, gu2024mamba} have emerged as an efficient, linear-time alternative to attention for sequence modeling.
To apply SSMs to images, Vision Mamba variants serialize 2D feature maps into 1D token sequences. Early designs explore bidirectional and cross-axis scanning to mitigate the loss of 2D spatial structure~\cite{zhu2024vimmamba, liu2024vmamba}, while subsequent works further optimize scan-order designs to better preserve locality and continuity~\cite{munir2025vcmamba,hatamizadeh2025mambavision}.

SSM-based approaches for VFI remain relatively underexplored.
VFIMamba~\cite{zhang2024vfimamba} adapts selective scanning to two-frame modeling by arranging tokens from $I_0$ and $I_1$ into an interleaved sequence for efficient multi-directional mixing.
LC-Mamba~\cite{jeong2025lc} emphasizes locality by introducing shifted-window and Hilbert curve-based scan orders, further extending scanning to spatiotemporal tokens.
While these methods demonstrate that SSMs can serve as highly efficient feature mixers for VFI, their scan orders remain defined over predetermined token traversals on static grids (spatial or spatiotemporal).

Unlike prior SSM-based VFI methods, we avoid serializing tokens along fixed spatial traversals. Instead, we sample features along flow-guided trajectories and use the resulting forward and backward contexts for both frame synthesis and intermediate motion refinement. This trajectory-aligned aggregation helps mitigate alignment errors in large-displacement and occlusion-heavy scenarios.

\section{Method}\label{sec:method}

\subsection{Overview}\label{sec:method_overview}
\begin{figure}[t]
\centering
\includegraphics[width=\linewidth]{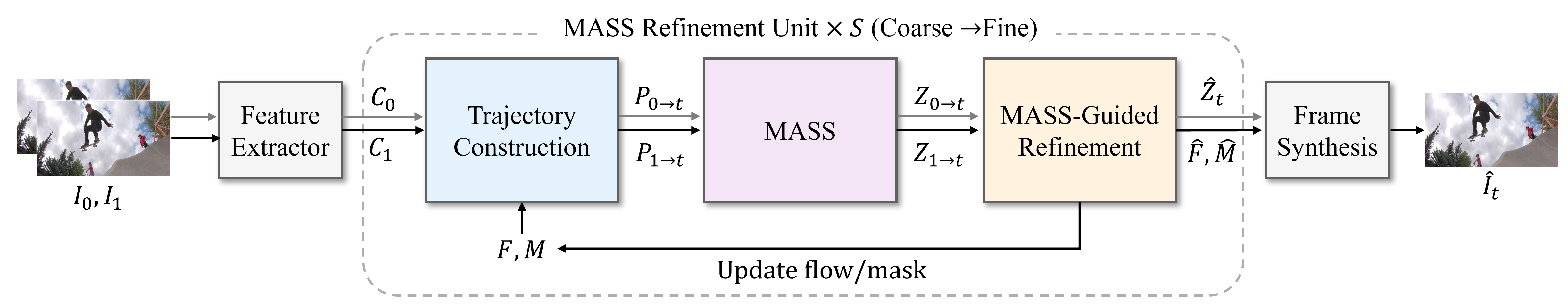}
\caption{Overall pipeline of our framework. Features extracted from $I_0$ and $I_1$ are processed through a cascade of MASS refinement units in a coarse-to-fine manner, producing refined motion and context to synthesize the target frame $\hat{I}_t$.}
\label{fig_over}
\end{figure}

Our goal is to synthesize an intermediate frame $\hat{I}_t$ at time $t \in (0,1)$ (default $t=0.5$) from two consecutive frames $I_0$ and $I_1$.
As illustrated in Fig.~\ref{fig_over}, the overall pipeline follows a coarse-to-fine design to robustly handle large non-linear motions and complex occlusions.
We first extract feature maps $C_0$ and $C_1$ from $I_0$ and $I_1$ using a shared feature extractor.
At each scale, the model repeats three steps: trajectory construction, trajectory-wise scan, and motion refinement.

Specifically, we start from coarse bidirectional flows $F=\{F_{t\to 0}, F_{t\to 1}\}$ (and an occlusion/blending mask $M$) estimated by a lightweight estimator~\cite{kong2022ifrnet}.
These coarse motions provide a baseline for building pixel-wise trajectories $P_{0\to t}$ and $P_{1\to t}$.
Unlike standard VFI methods that aggregate features on a static grid (e.g., via regular convolutions or window attention), our MASS aggregates features along these trajectories.
As a result, MASS produces forward and backward motion-aligned contexts $Z_{0\to t}$ and $Z_{1\to t}$, which are later fused into $\hat{Z}_t$ for frame synthesis.

Crucially, the aggregated features serve two roles in our framework.
First, they provide rich semantic context for frame synthesis. By gathering information along the motion trajectory, they preserve complex structures and fine textures that pure flow-based warping often misses. For synthesis, the forward and backward contexts are fused into $\hat{Z}_t$ and combined with the refined motion in the synthesis module to generate the target frame $\hat{I}_t$.

Second, the aggregated features provide a cue for motion reliability.
In non-occluded regions, forward and backward trajectories should trace the same physical content, so their aggregated contexts should be consistent. We use the discrepancy between these contexts in the refinement module to identify trajectory errors and occlusions, and to update the coarse flow and mask, producing refined estimates $\hat{F}$ and $\hat{M}$. These updated motion estimates are then fed into the next stage to reconstruct more accurate trajectories, forming an iterative coarse-to-fine update loop.

\subsection{Motion-Aligned Selective Scan}\label{sec:mass}
To overcome the limitations of static grid aggregation, MASS reformulates VFI as a trajectory-wise sequence modeling task. For a target pixel $p$, instead of restricting the receptive field to a spatial window, we utilize the initial coarse motion to trace a continuous flow-guided path. We then sequentially sample features along this trajectory and aggregate them using an SSM , explicitly aligning the context with physical motion and avoiding interference from misaligned neighbors.

This process involves two coupled mechanisms.
First, we construct non-linear trajectories from coarse motion estimates, allowing the sampling process to follow complex motion paths rather than relying on a straight-line assumption.
Second, we aggregate the resulting trajectory-aligned feature sequences using a velocity-aware selective SSM (VA-SSM) to obtain motion-aligned contexts.
We apply this process symmetrically to both forward ($0 \to t$) and backward ($1 \to t$) directions, producing $Z_{0\to t}$ and $Z_{1\to t}$.
These contexts are then used not only for synthesis but also to guide motion refinement, as summarized in Fig.~\ref{fig_detail}. 

\begin{figure}[t]
\centering
\includegraphics[width=\linewidth]{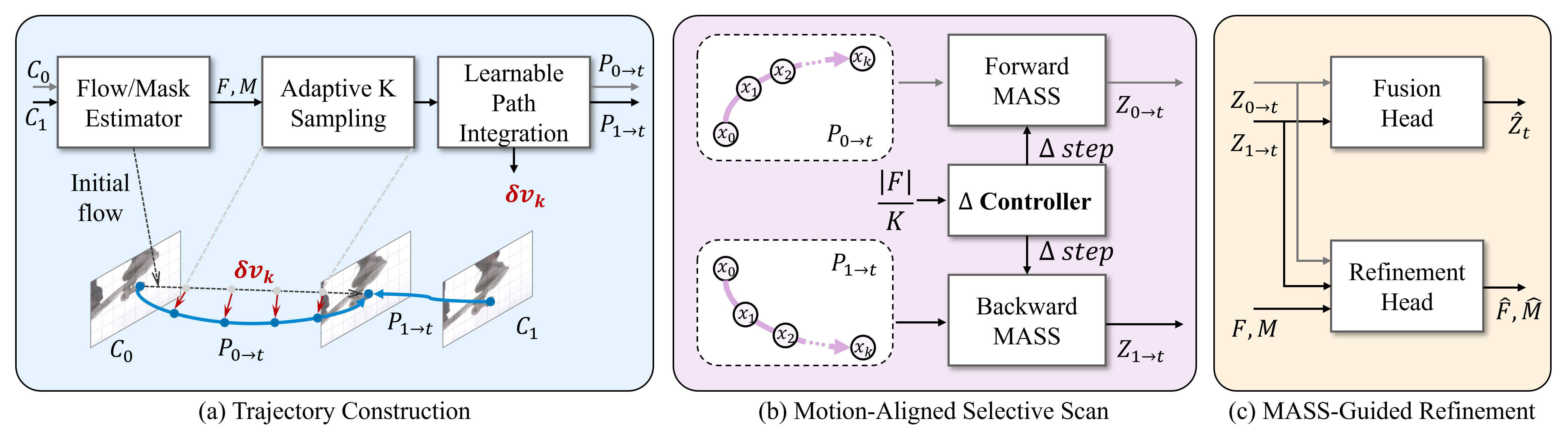}
\caption{Detailed illustration of the MASS refinement unit. (a) Non-linear trajectories are constructed from coarse flows through motion-adaptive sampling and learnable residual updates ($\delta \mathbf{v}_k$). (b) Trajectory-aligned feature sequences are aggregated with a velocity-aware scan, which adaptively modulates the discretization step according to the trajectory velocity ($|F|/K$). (c) The resulting forward and backward contexts are fused into $\hat{Z}_t$ for frame synthesis, while their discrepancy is used to refine the intermediate motion estimates ($\hat{F}, \hat{M}$).}
\label{fig_detail}
\end{figure}

\subsubsection{Learnable Non-Linear Path Integration.}\label{sec:mass_traj}
Linear interpolation assumes constant velocity, approximating motion as a straight line.
However, complex real-world dynamics (e.g., rotation, acceleration) require curved trajectories.
We construct such trajectories via an iterative integration process that progressively refines sampling coordinates starting from the coarse flow.

We first determine how densely to sample the trajectory.
We assign a dynamic sampling budget $K(p)$ based on the motion magnitude.
Intuitively, pixels with larger flow magnitude require denser sampling to avoid undersampling long trajectories.
If we sample these long paths with too few points, sparse sampling can under-approximate curved motion and may lead to aliasing artifacts.
Conversely, static regions require fewer steps.

Therefore, we dynamically set the sampling budget $K(p)$ by scaling the average magnitude of the bidirectional flows, defined as $\|F(p)\|=\frac{1}{2}(\|F_{t\to 0}(p)\|+\|F_{t\to 1}(p)\|)$. Using a scaling factor $\alpha$, we round the scaled magnitude to the nearest integer and restrict it to a predefined range $[K_{\min}, K_{\max}]$:
\begin{equation}\label{ep1}
K(p)=\text{clip}\left(\left\lfloor\alpha\cdot\|F(p)\|\right\rceil,K_{\min},K_{\max}\right)  
\end{equation}

Given the sampling budget $K(p)$, we define the trajectory by iteratively updating sampling coordinates relative to the fixed target pixel.
Let $\mathbf{u}$ denote the coordinate of pixel $p$ in the target frame.
We aim to trace a sequence of points $\{\mathbf{u}+\mathbf{d}_k\}_{k=0}^{K(p)}$ towards the source frame, so that sampling follows the underlying motion.
We initialize the accumulated displacement as zero ($\mathbf{d}_0 = \mathbf{0}$).
At each step $k$, we first sample a local feature vector from the source feature map at the current traced coordinate $\mathbf{u}+\mathbf{d}_k$ via the backward warping operation~\cite{kong2022ifrnet}, denoted as $\overleftarrow{\mathcal{W}}(C_0,\mathbf{u}+\mathbf{d}_k)$.
We then feed this motion-aligned feature together with the coarse flow $F_{t\to 0}(p)$ and the normalized progress indicator
$r_k = k/K(p)$ into two convolutional layers $\phi_{0\to t}$, which predict a residual velocity $\delta \mathbf{v}_k$.
This residual term serves as a learnable correction to the constant-velocity assumption; it lets the trajectory deviate from linear interpolation when local appearance indicates curved motion.
For the forward direction ($0\to t$), this is written as:
\begin{equation}
\delta \mathbf{v}_k = \phi_{0\to t}\!\left( \overleftarrow{\mathcal{W}}(C_0, \mathbf{u} + \mathbf{d}_k),\; F_{t\to 0}(p),\; r_k \right),
\quad \text{with } r_k=\frac{k}{K(p)}.
\end{equation}
The displacement is then updated by taking one coarse-flow step (a constant increment of $F_{t\to 0}(p)/K(p)$) and adding the predicted residual:
\begin{equation}
\mathbf{d}_{k+1} = \mathbf{d}_k + \frac{F_{t\to 0}(p)}{K(p)} + \delta \mathbf{v}_k.
\end{equation}
In other words, each step advances the trajectory by a coarse-flow increment plus a learned residual offset, allowing the path to bend when local appearance deviates from the straight-line assumption.
Repeating this process yields a non-linear trajectory and a trajectory-aligned feature sequence.
We apply the same construction to the backward direction ($1\to t$) using $C_1$ and $F_{t\to 1}$, producing $P_{1\to t}$.
To enable efficient batched computation, we implement variable-length trajectories using a fixed-length buffer with a masking strategy.
More details are provided in the supplementary material.

\subsubsection{Velocity-Aware Selective Scan.}\label{sec:mass_scan}

Given the extracted feature sequence, the subsequent step is to aggregate these observations into a unified context vector $Z(p)$. Rather than relying on a aggregation mechanism, we introduce a Velocity-Aware SSM (VA-SSM) to adaptively modulate the internal state transition dynamics based on the underlying per-step motion along the trajectory.
Intuitively, larger displacements per step correspond to faster appearance changes along the sequence.
A fixed discretization may under-resolve rapid appearance variation along high-velocity trajectories, which can reduce the fidelity of fine textures.

To address this, we modulate the discretization step size according to the trajectory's average velocity. In discretized SSMs, the system evolution depends on a step size $\Delta$. While $K(p)$ controls how densely we sample points in space along the trajectory, $\Delta$ controls how finely the SSM evolves along the resulting token sequence.
Since the trajectory spans an approximate displacement of $\|F(p)\|$ with $K(p)$ steps in total, we define the average trajectory velocity as $v(p) = \|F(p)\| / K(p)$, where 
$\|F(p)\|$ is the average flow magnitude introduced above. Because $K(p)$ is discretized to a bounded range, the resulting step velocity still varies across pixels. We then adjust the trajectory-specific step size $\Delta(p)$ inversely with $v(p)$, which changes the effective transition used in the scan:

\begin{equation}\label{ep4}
    \overline{A}(p) = \exp(\Delta(p)A), \quad \text{with} \quad \Delta(p) = \frac{1}{1 + \beta \cdot v(p)}, \quad v(p) = \frac{\|F(p)\|}{K(p)}
\end{equation}

By reducing $\Delta(p)$ for trajectories with large average velocities, the model updates its internal state with finer temporal granularity, preserving high-frequency details such as edges and textures during context aggregation.
Applying the selective scan on the forward and backward trajectories yields the motion-aligned contexts $Z_{0\to t}$ and $Z_{1\to t}$.

\subsection{MASS-Guided Refinement and Synthesis.}\label{sec:Refinement}
The MASS operation yields two motion-aligned contexts, $Z_{0\to t}$ and $Z_{1\to t}$. We utilize these features simultaneously to correct geometric errors and to synthesize the final frame.

We first identify trajectory errors by checking the consistency between the forward and backward contexts. Ideally, for non-occluded regions, the semantic context aggregated along the underlying motion path should be similar regardless of the scanning direction. We compute a discrepancy map $D(p) = |Z_{0\to t} - Z_{1\to t}|$, where a large difference indicates trajectory misalignment or occlusion. We use the discrepancy between forward and backward contexts as a refinement cue. A two convolutional layer head takes this cue, together with the aggregated contexts and the current motion estimates, and predicts residual updates to produce the refined flow $\hat{F}$ and mask $\hat{M}$. In this way, visual consistency directly guides geometric correction, and the refined motion is reused in the subsequent iteration to improve trajectory construction.

Meanwhile, frame synthesis requires a unified semantic representation. A two convolutional layer fusion head merges $Z_{0\to t}$ and $Z_{1\to t}$ into a single context $\hat{Z}_t$. This head employs a softmax-based gating operation to effectively down-weight unreliable information from occluded regions. Using the refined motion and fused context, we generate the target frame by combining motion-compensated warping with residual synthesis. We first compute a structural baseline $I_{\text{base}}$ by warping the input frames using the refined flow $\hat{F}$ and mask $\hat{M}$:

\begin{equation}
I_{\text{base}} = \hat{M} \odot \overleftarrow{\mathcal{W}}(I_0, \hat{F}_{t\to 0}) + (1-\hat{M}) \odot \overleftarrow{\mathcal{W}}(I_1, \hat{F}_{t\to 1}).
\end{equation}
Here, $\odot$ represents the Hadamard product.
Subsequently, the generator $\mathcal{G}$ synthesizes the final image by integrating the global motion context $\hat{Z}_t$ with the local source features ($C_0, C_1$) to restore fine-grained details:
\begin{equation}
\hat{I}_t = I_{\text{base}} + \mathcal{G}(\hat{Z}_t, C_0, C_1).
\end{equation}
In this formulation, $I_{\text{base}}$ provides the structural foundation based on corrected motion, handling large displacements.
For the generator $\mathcal{G}$, we adopt a U-Net-shaped architecture~\cite{huang2022real}. By feeding both the aligned MASS context and raw local features into this architecture, $\mathcal{G}$ effectively recovers complex textures and edges that simple warping might miss. 
More details are provided in the supplementary material.

\section{Experimental Results}
\label{sec:experiments}

\subsection{Datasets}\label{sec:datasets}
We evaluate our proposed method on comprehensive benchmark datasets covering a wide range of resolutions and motion complexities.
\textbf{Vimeo90K}~\cite{xue2019video} is our primary training set. It contains 51,312 triplets for training and 3,782 frame triplets for testing, with a resolution of $448 \times 256$. This dataset is widely recognized for its diversity in scene content and realistic motion patterns.
\textbf{UCF101}~\cite{soomro2012ucf101} includes a test set of 379 frame triplets at $256 \times 256$. 
\textbf{Middlebury}~\cite{baker2011database} uses the OTHER set at about $640 \times 480$ resolution. 
\textbf{SNU-FILM}~\cite{choi2020channel} includes 1,240 frame triplets at $1280 \times 720$, categorized by motion difficulty (Easy, Medium, Hard, Extreme). 
Lastly, \textbf{Xiph}~\cite{xiph1994} is tested in two versions to evaluate high-resolution performance: Xiph-2K, with images down-sampled to 2K, and Xiph-4K, with centrally cropped 4K images.

\subsection{Implementation Details}
\label{subsec:implementation}
Our framework is implemented in PyTorch. We train our model using the AdamW optimizer with a weight decay of $1 \times 10^{-4}$. The training is conducted on two NVIDIA RTX 3090 GPUs with a total batch size of 32.
The feature extractor adopts the SGM-VFI~\cite{liu2024sparse} architecture, where the deepest features are at $H/8 \times W/8$ resolution. The refinement is performed progressively, with the $1/8$ and $1/4$ scales serving as the coarse and fine stages, respectively, via upsampling.
For the MASS module, we set the dynamic sampling budget range $[K_{\min}, K_{\max}]$ to $[2, 8]$.
The discretization step size $\Delta$ for the Velocity-Aware SSM is adaptively modulated within the range of $[0.25, 1.0]$ based on motion magnitude.
The scaling parameters $\alpha$ and $\beta$ in Eq.~\ref{ep1} and Eq.~\ref{ep4} are dynamically computed by normalizing local motion magnitude and velocity against their frame-level averages. 
For the SSM configuration, we balance efficiency and performance by setting the internal state dimension to $16$ and the expansion factor to $2$.
For the objective function, we utilize the Laplacian pyramid loss~\cite{niklaus2020softmax} to guide the network in reconstructing multi-scale structures and fine details.
Further details are provided in the supplementary material.
 
\begin{table}[t]\caption{Quantitative comparison across benchmarks (IE for Middlebury (M.B.); PSNR/SSIM for Vimeo90K, UCF101, Xiph, and SNU-FILM). The best and
second-best results are in \textbf{bold} and \underline{underlined}, respectively.  $\dagger$ indicates models originally trained with additional datasets, which we re-trained exclusively on Vimeo90K for fair comparison. Other results are derived from~\cite{hu2022many, huang2022real, kong2022ifrnet, lu2022video, reda2022film, li2023amt, jeong2025lc}. Evaluation procedures follow~\cite{soomro2012ucf101} for Vimeo90K, UCF101, and Middlebury, \cite{niklaus2020softmax} for Xiph, and \cite{kong2022ifrnet} for SNU-FILM, with test-time augmentation disabled. FLOPs are measured at a resolution of 1280$\times$720.} \label{tab:quantitative}
\centering
\resizebox{\linewidth}{!}{%
\begin{tabular}{lcccccccccccc}

\toprule
\multirow{2}{*}{Method}  & \multirow{2}{*}{M.B.} & \multirow{2}{*}{Vimeo90K} & \multirow{2}{*}{UCF101} & \multicolumn{2}{c}{Xiph} & \multicolumn{4}{c}{SNU-FILM} & \multirow{2}{*}{FLOPs (T)} \\ \cmidrule(lr){5-6} \cmidrule(lr){7-10}
  & & &  & 2K & 4K &Easy & Medium & Hard & Extreme &   \\ \midrule

IFRNet-B~\cite{kong2022ifrnet}& 1.95 & 35.80/0.9794 & 35.29/0.9693 & 36.00/0.936 & 33.99/0.893  & {40.03}/0.9905 & 35.94/0.9793 & 30.41/0.9358 & 25.05/0.8587 & 0.21  \\

RIFE~\cite{huang2022real} & 1.96 &35.61/0.9779 & 35.28/0.9690 & 36.19/0.938 & 33.76/0.894 &  39.80/0.9903 & 35.76/0.9787 & 30.36/0.9351 & 25.27/0.8601  & 0.20\\

AdaCoF~\cite{lee2020adacof} & 2.24 &34.47/0.9730 & 34.90/0.9680 & 34.86/0.928 & 31.68/0.870 &  39.80/0.9900 & 35.05/0.9754 & 29.46/0.9244 & 24.31/0.8439  & 0.36  \\


ABME~\cite{park2021asymmetric} & 2.01 & 36.18/0.9805 & 35.38/0.9698 & 36.53/0.944 & 33.73/0.901 & 39.59/0.9901 & 35.77/0.9789 & 30.58/0.9364 & 25.42/0.8639 &  1.30 \\

DAIN~\cite{bao2019depth}  & 2.04 &34.71/0.9756 & 34.99/0.9683 & 35.95/0.940 & 33.49/0.895 &  39.73/0.9902 & 35.46/0.9780 & 30.17/0.9335 & 25.09/0.8584 & 5.51 \\

VFIFormer~\cite{lu2022video}& \underline{1.82} & {36.50}/{0.9815} & 35.42/0.9699 & 36.89/0.945 & 34.50/0.906  & 40.12/0.9907 & 36.09/0.9798 & 30.67/0.9378 & 25.43/0.8643 & 47.71  \\

AMT-L~\cite{li2023amt} &1.86  & {36.34}/{0.9813} & 35.39/0.9697 & 36.27/0.940 & {34.48}/0.903 & 39.95/0.9904 & {36.09}/0.9797 & {30.75}/0.9377 & 25.41/0.8633 & 0.68  \\

AMT-G~\cite{li2023amt} & 1.83  & \textbf{36.53}/\textbf{0.9816}& 35.41/0.9698 & 36.38/0.940 & 34.63/0.903 & 39.88/\textbf{0.9912} & 36.12/\textbf{0.9805} & 30.77/0.9384 & 25.42/0.8644 & 2.2 \\

EMA-S~\cite{zhang2023extracting}& 1.94 & 36.07/0.9794 & 35.34/0.9696 & 36.54/0.942 & 34.24/0.902  & 39.81/0.9903 & 35.88/0.9792 & 30.68/0.9371 & 25.47/0.8627 & 0.39  \\

EMA~\cite{zhang2023extracting}  & 1.84 & 36.50/0.9814 & 35.38/0.9697 & 36.74/0.944 & 34.54/0.905 & 39.57/0.9905 & 35.85/0.9797 & 30.80/\underline{0.9389} & 25.59/0.8650 & 1.64 \\

SGM-VFI-S-1/2$^\dagger$~\cite{liu2024sparse} & 1.96 & 35.98/0.9796 & 35.32/0.9695& 36.58/0.942 & 34.23/0.901  &  40.06/0.9908 & 36.02/0.9796 & 30.58/0.9364  & 25.25/0.8590&  1.96 \\

VFIMamba-S$^\dagger$~\cite{zhang2024vfimamba} & 1.98  & 35.87/0.9793 & 35.32/0.9697 & 36.26/0.940  & 33.52/0.890  &  40.00/0.9907   & 35.81/0.9791  & 30.58/{0.9378} & 25.30/0.8620 & 0.39  \\

VFIMamba$^\dagger$~\cite{zhang2024vfimamba}  & 1.90 & 36.46/0.9814 & 35.42/\underline{0.9700} & 36.82/0.944 & 34.12/0.898 &  \textbf{40.20}/\underline{0.9911} & 36.11/0.9800 & 30.79/\underline{0.9389}  & 25.66/0.8657  &  1.54\\

LC-Mamba-B~\cite{jeong2025lc} & {1.89} &  {36.43}/{0.9813} & {35.39}/{0.9698} & {36.90/0.945} & 34.26/{0.904} & {40.07/0.9909} & 36.08/{0.9801} & 30.59/0.9375 & 25.35/{0.8630} & 1.06  \\

LC-Mamba-P~\cite{jeong2025lc} &  1.92 & \textbf{36.53}/\textbf{0.9816} & 35.42/0.9699 & \underline{36.99}/\textbf{0.946} & 34.49/0.906 &   {40.16}/0.9909 & \underline{36.17}/\underline{0.9802} & 30.72/0.9382 & 25.48/0.8645 & 2.70\\

\midrule

Ours-S &{1.90} & 36.32/0.9805 & \underline{35.44}/0.9699 & {36.84}/{0.944} & {34.65/0.907}& 40.05/0.9908 & {36.10}/{0.9800} & \underline{30.90}/0.9387 & \underline{25.73}/\underline{0.8668} &  0.50  \\

Ours & \textbf{1.79} & \textbf{36.53}/\textbf{0.9816} & \textbf{35.48}/\textbf{0.9703} & \textbf{37.05}/\textbf{0.946} & \textbf{34.83}/\textbf{0.908}   & \underline{40.17}/0.9909 & \textbf{36.23}/\underline{0.9802} & \textbf{31.09}/\textbf{0.9407} & \textbf{26.04}/\textbf{0.8681} & 2.06 \\

\bottomrule
\end{tabular}%
}
\end{table}

\begin{figure}[t]
\centering
\includegraphics[width=.85\linewidth]{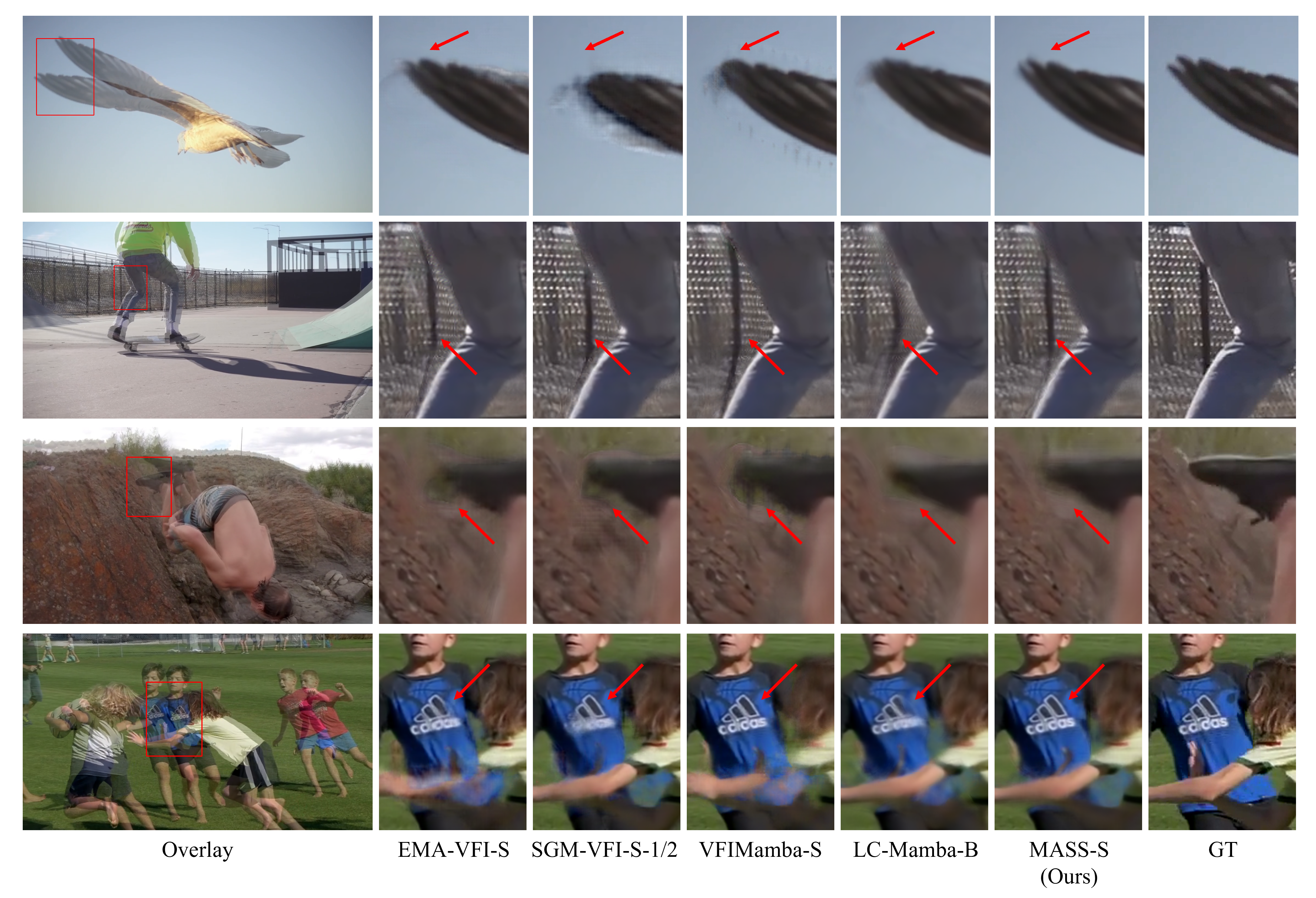}
\caption{Visual comparison on SNU-FILM~\cite{choi2020channel}. The red arrows highlight regions with large motions and fine details.}
\label{fig:VOT}
\end{figure}

\subsection{Comparison with Previous Methods} \label{subsec:comparison}  
To evaluate the proposed MASS framework, we compare it against recent VFI methods across five standard benchmarks, as summarized in Table~\ref{tab:quantitative}. 
For a fair comparison, we evaluate all models under the same training data constraint. Since some recent methods utilize additional datasets (e.g., X4K1000FPS~\cite{sim2021xvfi}) for fine-tuning, we re-trained them from scratch exclusively on the standard Vimeo90K dataset using their official codebases (denoted by $\dagger$).
More detailed analyses are provided in the supplementary material.

While MASS delivers highly competitive performance on standard benchmarks such as Vimeo90K and UCF101, it shows the clearest gains in challenging scenarios characterized by large, non-linear motions and complex occlusions. This is particularly noticeable on the SNU-FILM dataset, which categorizes videos by motion magnitude. As motion complexity increases, performance degrades as large motion and occlusion make alignment increasingly difficult. In contrast, MASS demonstrates clear performance improvements on both the Hard and Extreme subsets. Notably, on the Extreme split, our method yields 26.04 dB in PSNR, outperforming the previous best approach, VFIMamba~\cite{zhang2024vfimamba}, by 0.38 dB. This gain is consistent with the benefit of modeling non-linear, trajectory-aligned context under complex motion.

Furthermore, the proposed model maintains consistent performance in high-resolution scenarios. On the Xiph-2K and Xiph-4K benchmarks, MASS compares favorably against all competing methods, recording 37.05 dB and 34.83 dB in PSNR, respectively, outperforming recent state-of-the-art methods such as LC-Mamba. These results suggest that the velocity-aware scan helps preserve details in demanding high-resolution, large-displacement scenarios. Additionally, our lightweight variant, MASS-S, maintains competitive restoration quality with a reduced computational cost of 0.50 TFLOPs, demonstrating its applicability in resource-constrained environments.

Finally, visual comparisons in Fig.~\ref{fig:VOT} support the quantitative results. While prior methods frequently exhibit ghosting or severe blurring under large displacements, MASS reconstructs structurally coherent frames with sharper edges. This highlights the advantage of our trajectory-wise modeling. By tracking non-linear motion paths, MASS explicitly ensures semantic continuity, successfully mitigating the feature misalignment that typically degrades performance in dynamic scenes.

\begin{table}[t]
    \centering
    \renewcommand{\tabcolsep}{2mm}
    \caption{
    Ablation study on different configurations of the MASS module and integrator variants. Performance is reported in PSNR/SSIM.}
    \label{tab:ablation_mass_plus_integrator}
    \resizebox{\linewidth}{!}{
    
        \begin{tabular}{lcccccccc}
            \toprule
            \multirow{2}{*}{Settings} &
            \multirow{2}{*}{\shortstack{Learnable\\Path}} &
            \multirow{2}{*}{\shortstack{Adpt.\\$K$}} &
            \multirow{2}{*}{\shortstack{Vel. $\Delta$}} &
            \multicolumn{2}{c}{SNU-FILM} &
            \multicolumn{2}{c}{Xiph} &
            \multirow{2}{*}{FLOPs (T)} \\
            \cmidrule(lr){5-6} \cmidrule(lr){7-8}
             & & & & Hard & Extreme & 2K & 4K & \\
            \midrule

            \multicolumn{9}{l}{\textbf{MASS components}} \\
            \midrule
            Baseline (Linear + Vanilla SSM) &
            -- & -- & -- &
            30.66/0.937 & 25.49/0.863 & 36.54/0.941 & 34.40/0.904 & 0.53 \\

            + Non-linear Traj. &
            \checkmark & -- & -- &
            30.81/0.938 & 25.60/0.865 & 36.69/0.942 & 34.54/0.905 & 0.54 \\

            + Adpt. Sampling &
            \checkmark & \checkmark & -- &
            30.85/0.938 & 25.65/0.865 & 36.76/0.943 & 34.58/0.905 & 0.49 \\

            \textbf{+ MASS (Ours)} &
            \checkmark & \checkmark & \checkmark  &
            \textbf{30.90}/\textbf{0.939} & \textbf{25.73}/\textbf{0.867} &
            \textbf{36.84}/\textbf{0.944} & \textbf{34.65}/\textbf{0.907} & 0.50 \\

            \midrule
            \multicolumn{9}{l}{\textbf{Integrator component}} \\
            \midrule
            1D Conv &
            \checkmark & \checkmark & -- &
            30.68/0.938 & 25.53/0.864 & 36.39/0.941 & 34.07/0.900 & 0.50 \\

            LSTM &
            \checkmark & \checkmark & -- &
            30.80/0.938 & 25.56/0.865 & 36.30/0.940 & 34.10/0.900 & 0.54 \\

            Self-Attention &
            \checkmark & \checkmark & -- &
            30.83/0.938 & 25.68/0.866 & 36.60/0.938 & 34.20/0.902 & 0.52 \\
            
            \textbf{VA-SSM (Ours)} &
            \checkmark & \checkmark & \checkmark  &
            \textbf{30.90}/\textbf{0.939} & \textbf{25.73}/\textbf{0.867} &
            \textbf{36.84}/\textbf{0.944} & \textbf{34.65}/\textbf{0.907} & 0.50 \\
            \bottomrule
        \end{tabular}
    }
\end{table}
\subsection{Ablation Study}
\label{subsec:ablation}
To validate the effectiveness of the proposed MASS framework, we conduct component-wise ablation studies using the lightweight variant of our model (Ours-S). We report PSNR and SSIM on the SNU-FILM and Xiph datasets, alongside FLOPs measured at 720p resolution, computed using the mean adaptive $K$ on the SNU-FILM Hard split. Additional ablation results are provided in the supplementary material.

\subsubsection{MASS Components and Integrator Variants.}
Table~\ref{tab:ablation_mass_plus_integrator} investigates the contribution of each component within our MASS framework, alongside an ablation of the integrator design. Since MASS constructs token sequences along motion trajectories rather than a predefined spatial order, directly replacing the trajectory construction with a static grid scan is not straightforward within the same architecture. To validate the effectiveness of our motion-aligned design, the upper part of Table~\ref{tab:ablation_mass_plus_integrator} begins with a trajectory-based baseline using a linear path, a fixed sampling budget ($K(p)=8$), and a vanilla SSM, and then incrementally adds each proposed component. Transitioning to a learnable non-linear path yields consistent gains, as learnable residual offsets effectively model complex, continuous motion dynamics. Furthermore, our adaptive sampling strategy dynamically allocates points based on motion magnitude. Beyond lowering FLOPs, it yields higher accuracy than using a fixed maximum sampling budget, suggesting that sequence simplification is actively beneficial for static or small motions. Finally, integrating the velocity-aware step size (Vel. $\Delta$) converts the baseline SSM into our VA-SSM. By dynamically modulating the discretization step, the model adjusts its internal state transitions based on the trajectory velocity. This suggests that fast-moving trajectories are aggregated with appropriately fine granularity, while static or slow-moving regions maintain stable state representations.

Next, we evaluate the integrator component using the optimized trajectory sampling. As shown in the lower section of Table~\ref{tab:ablation_mass_plus_integrator}, standard sequence models such as 1D convolutions, LSTMs, and self-attention yield sub-optimal performance or incur higher computational costs. In contrast, our VA-SSM adaptively modulates its internal state transitions based on trajectory velocity. This dynamic aggregation achieves the highest accuracy across all benchmarks while matching the low FLOPs of a simple 1D convolution, demonstrating an optimal accuracy-efficiency balance uniquely suited for trajectory-wise modeling.

\subsubsection{Scale Iterations and MASS-Guided Refinement.}
Table~\ref{tab:ablation_scale_and_heads} analyzes the trade-off between restoration accuracy and computational efficiency regarding the number of scale iterations ($S$). A single MASS refinement unit ($S=1$) provides insufficient refinement and feature aggregation, resulting in limited performance. Increasing the iterations to $S=2$ significantly improves accuracy, suggesting that iterative processing enables finer refinement and boosts performance in scenarios with complex motion. However, adding a third iteration ($S=3$) yields only marginal performance gains while substantially increasing the computational overhead. Therefore, we adopt $S=2$ as the default configuration to maximize performance while maintaining practical efficiency.

In the lower section of Table~\ref{tab:ablation_scale_and_heads}, we further validate the necessity of the MASS-guided fusion and refinement heads. First, when the fusion head is removed, we observe that a naive aggregation of MASS features (e.g., simple element-wise addition) is insufficient. This result shows that our gating mechanism is necessary to effectively integrate the results from bidirectional contexts. Furthermore, the absence of the refinement head leads to a significant performance degradation. This suggests that leveraging the discrepancies between bidirectional contexts is essential for correcting geometric errors in the intermediate flow and occlusion mask.

\begin{table}[t]
    \centering
    \renewcommand{\tabcolsep}{3.8mm}
    \caption{Ablation study on the number of scale iterations and the MASS-guided refinement heads. Performance is reported in PSNR/SSIM.}
    \label{tab:ablation_scale_and_heads}
    \resizebox{\linewidth}{!}{
        \begin{tabular}{lcccccc}
            \toprule
            \multirow{2}{*}{Settings} &
            \multicolumn{2}{c}{SNU-FILM} &
            \multicolumn{2}{c}{Xiph} &
            \multirow{2}{*}{Params (M)} &
            \multirow{2}{*}{FLOPs (T)} \\
            \cmidrule(lr){2-3} \cmidrule(lr){4-5}
             & Hard & Extreme & 2K & 4K & & \\
            \midrule

            \multicolumn{7}{l}{\textbf{Scale iterations}} \\
            \midrule
            $S=1$ & 30.82/0.938 & 25.64/0.864 & 36.60/0.942 & 34.40/0.905 & 15.7 & 0.45 \\
            $\mathbf{S=2}$ \textbf{(Ours)} & 30.90/0.938 & 25.73/0.866 & 36.84/0.944 & 34.65/0.907 & 16.5 & 0.50 \\
            $S=3$ & \textbf{30.92/0.939} & \textbf{25.75/0.867} & 36.84/0.944 & \textbf{34.69}/0.907 & 17.2 & 0.68 \\

            \midrule
            \multicolumn{7}{l}{\textbf{MASS-guided refinement}} \\
            \midrule
            w/o Fusion Head & 30.63/0.937 & 25.50/0.860 & 36.49/0.941 & 34.26/0.904 & 16.4 & 0.49 \\
            w/o Refinement Head & 30.56/0.936 & 25.44/0.864 & 36.46/0.940 & 34.20/0.903 & 16.4 & 0.49 \\
            \textbf{Full Model (Ours)} & \textbf{30.90}/\textbf{0.938} & \textbf{25.73}/\textbf{0.866} & \textbf{36.84}/\textbf{0.944} & \textbf{34.65}/\textbf{0.907} & 16.5 & 0.50 \\
            \bottomrule
        \end{tabular}
    }
\end{table}

\subsection{In-Depth Analysis}

\subsubsection{Motion-Aligned Trajectory Controls.}
To examine the role of motion-aligned trajectory serialization in MASS, we conduct controlled experiments (Table~\ref{tab:mass_motion_alignment_controls}) while keeping network weights and inference settings fixed.We first evaluate the impact of trajectory direction. The default aligned trajectory consistently achieves the best performance. Rotating the flow by $\pm90^{\circ}$ (perpendicular to the motion) causes the most significant performance drop. This suggests that the model benefits from gathering context along the actual motion path rather than merely expanding the spatial sampling area. Negating or swapping flows results in smaller drops, likely because these paths still align with the underlying motion axis. Furthermore, altering the token serialization order via random permutation or reversal degrades quality, particularly on high-resolution sequences. Replacing the trajectory-guided scan with a static $3 \times 3$ spatial grid also leads to a large performance drop. This indicates that MASS relies on the sequential, motion-consistent ordering of features rather than just an unordered set of local points. We also analyze the role of residual-velocity integration ($\delta v_k$). While disabling it causes a modest performance drop, the gap between aligned and perpendicular trajectories persists. This implies that motion alignment itself is a key factor in the observed improvements.

Finally, Fig.~\ref{fig_detail} visualizes the learned trajectory correction. The overlays show how residual velocity updates bend the initial linear flow, allowing the sampling path to flexibly follow complex motions such as acceleration, rotation, and boundary movements.

\begin{table}[t]
    \centering
    \renewcommand{\tabcolsep}{9mm}
    \caption{Controlled perturbations of motion alignment in MASS.
    (A) the trajectory direction, (B) the serialization order, and (C) the residual-velocity integration while keeping all other settings fixed.}
    \label{tab:mass_motion_alignment_controls}
    \resizebox{\linewidth}{!}{
        \begin{tabular}{lcccc}
            \toprule
            \multirow{2}{*}{Settings} & 
            \multicolumn{2}{c}{SNU-FILM} &
            \multicolumn{2}{c}{Xiph} 
            \\
            \cmidrule(lr){2-3} \cmidrule(lr){4-5}
            & Hard & Extreme & 2K & 4K \\
            \midrule
            \multicolumn{5}{l}{\textbf{(A) Trajectory direction perturbation}} \\
            \midrule
            Aligned (default) & \textbf{30.90}/\textbf{0.939} & \textbf{25.73}/\textbf{0.867} &
            \textbf{36.84}/\textbf{0.944} & \textbf{34.65}/\textbf{0.907}
             \\
            Rotated $+90^{\circ}$ (perpendicular) & 30.61/0.937 & 25.52/0.866 & 36.44/0.941 & 34.34/0.902 \\
            Rotated $-90^{\circ}$ (perpendicular) & 30.63/0.937 & 25.51/0.866 & 36.45/0.941 & 34.38/0.903\\
            Negated flow ($-F_{t\rightarrow{0}}$, $-F_{t\rightarrow{1}}$) & 30.69/0.938 & 25.55/0.866 & 36.54/0.942 & 34.40/0.903 \\
            Swapped flows ($F_{t\rightarrow{1}}$, $F_{t\rightarrow{0}}$) & 30.68/0.938 & 25.54/0.866 & 36.52/0.942 & 34.39/0.903 \\
            \midrule   
            \multicolumn{5}{l}{\textbf{(B) Serialization order perturbation}} \\
            \midrule
            Trajectory order (default) & \textbf{30.90}/\textbf{0.939} & \textbf{25.73}/\textbf{0.867} &
            \textbf{36.84}/\textbf{0.944} & \textbf{34.65}/\textbf{0.907} \\    
            Random permutation & 29.77/0.927& 25.43/0.865 & 34.86/0.928 &32.91/0.880\\
            Reverse order & 29.30/0.921 & 25.23/0.860 & 33.44/0.916 & 32.02/0.872 \\

            Static grid order $(3\times3)$  &26.75/0.885 & 24.06/0.837 & 28.92/0.826 & 29.44/0.855\\
            \midrule
            \multicolumn{5}{l}{\textbf{(C) Residual-velocity integration ($\delta v_k$)}} \\
            \midrule
            Aligned ($\delta v_k$ ON) & \textbf{30.90}/\textbf{0.939} & \textbf{25.73}/\textbf{0.867} &
            \textbf{36.84}/\textbf{0.944} & \textbf{34.65}/\textbf{0.907} \\
            Aligned ($\delta v_k$ OFF) & 30.86/0.938 & 25.69/0.866 & 36.76/0.943 & 34.59/0.905\\
            Rotated +90° ($\delta v_k$ ON) & 30.61/0.937 & 25.52/0.866 & 36.44/0.941 & 34.34/0.902 \\
            Rotated +90° ($\delta v_k$ OFF) & 30.50/0.937 & 25.44/0.865 & 36.34/0.940 & 34.30/0.902\\
            \bottomrule
        \end{tabular}
    }
\end{table}

\begin{figure}[t]
\centering
\includegraphics[width=0.95\linewidth]{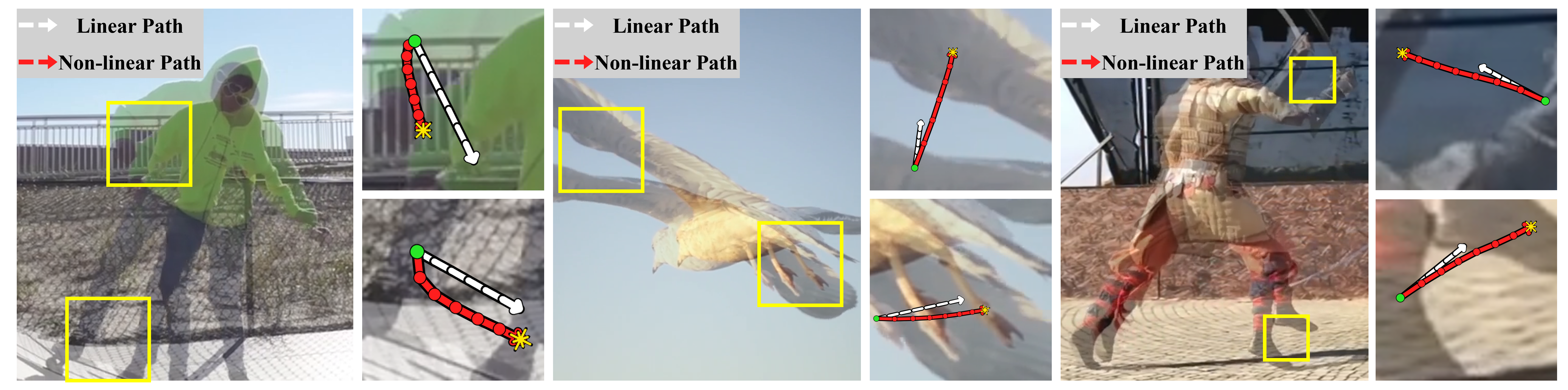}
\caption{ Visualization of the learned trajectory correction.}
\label{fig_detail}
\end{figure}

\subsubsection{MASS-Guided Refinement.}
\begin{figure}[t]
\centering
\includegraphics[width=.95\linewidth]{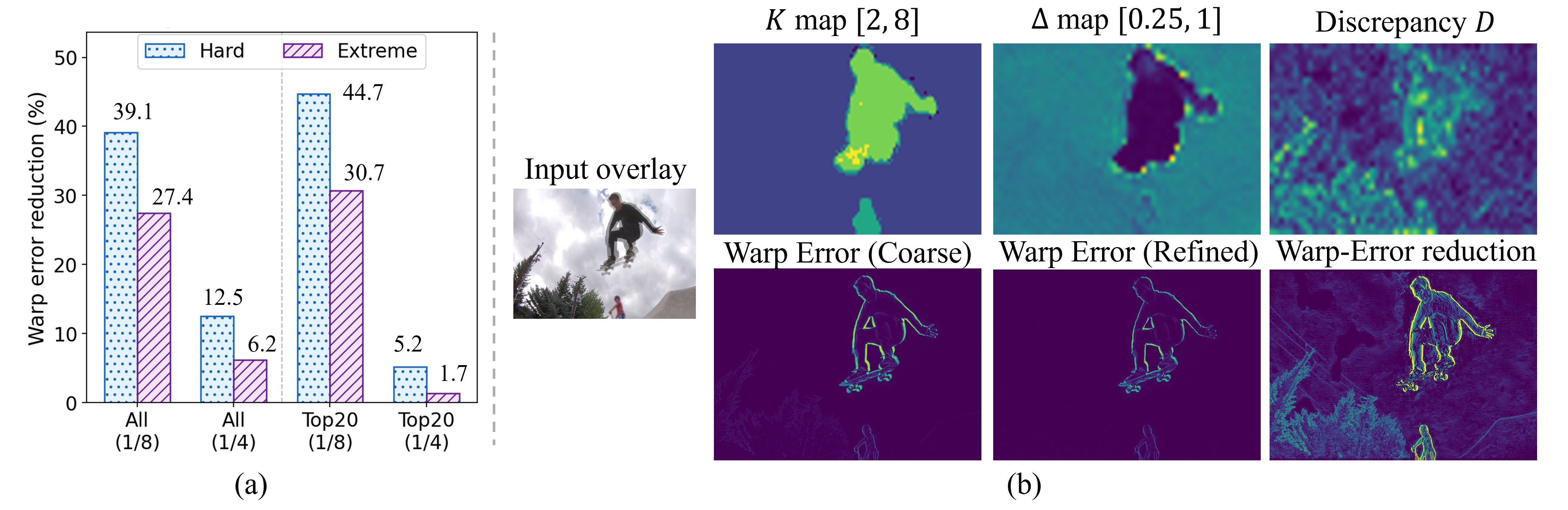}
\caption{In-depth analysis of MASS-guided refinement. (a) Average warp error reduction at the coarse $1/8$ scale and the fine $1/4$ scale for all pixels versus the top 20\% high-motion regions on the SNU-FILM~\cite{choi2020channel} Hard and Extreme splits. (b) Visualizations of internal representations at the coarse 1/8 scale: The input overlay is shown for spatial reference. }
    \label{fig:qualitative_analysis}
\end{figure}

Fig.~\ref{fig:qualitative_analysis} illustrates the internal mechanisms of MASS-guided refinement. Quantitatively, as shown in Fig.~\ref{fig:qualitative_analysis}(a), the refinement consistently reduces initial coarse warping errors across progressive feature scales (from the coarse 1/8 scale to the fine 1/4 scale). The most substantial corrections occur at the 1/8 scale, where severe initial motion errors are resolved, while finer geometric adjustments are left to the 1/4 scale. To further investigate performance in regions with complex motion, we further isolate the top 20\% of pixels with the highest motion magnitude. In these challenging regions, the error reduction is greater at the 1/8 scale, whereas it drops at the 1/4 scale. These results indicate that the 1/8 scale effectively resolves the majority of large-displacement errors, leaving less residual misalignment for the 1/4 scale to correct.

Qualitatively, Fig.~\ref{fig:qualitative_analysis}(b) visualizes this refinement process at the coarse 1/8 scale. A larger sampling budget $K$ is allocated to the fast-moving foreground subject based on its motion magnitude. Correspondingly, the velocity-aware step size $\Delta$ is adjusted to smaller values in these regions, enabling finer state updates along the dynamic path. The discrepancy map $D$ highlights local geometric misalignments. Guided by this map, the model refines the initial coarse motion, resulting in reduced errors compared to the coarse warping result. For a better comparison between the two warp error maps, we visualize the error reduction. The observed reduction suggests that the overall MASS mechanism effectively handles dynamic motions, leading to notable performance improvements.

\section{Conclusion}

In this paper, we presented MASS, a novel framework that redefines the application of SSMs for VFI. By shifting the feature scanning paradigm from static spatial grids to dynamic motion trajectories, MASS effectively captures complex, non-linear motion dynamics that conventional serialization fails to resolve. Our key innovations—learnable non-linear path integration and velocity-aware adaptive sampling—efficiently allocate computational resources while preserving high-frequency details in fast-moving regions. Moreover, by exploiting forward/backward trajectory consistency, MASS refines intermediate flows and occlusion masks in a robust coarse-to-fine manner. Extensive experiments demonstrate that MASS establishes new state-of-the-art results, exhibiting particular superiority in challenging scenarios with large displacements and complex dynamics. Despite its robustness, MASS inherently relies on the guidance of the initial coarse flow. In extreme cases where this initial estimation fails catastrophically (e.g., due to chaotic motion of tiny objects or severe motion blur), the trajectory-guided scanning may struggle to recover the correct semantic context. Future work will address this dependency by exploring joint optimization strategies, training the flow estimator and the SSM in a fully coupled manner to allow the SSM to correct flow guidance dynamically from the earliest stages.

\section*{Acknowledgements}
This work was supported by the National Reasearch Foundation of Koread (NRF) grant funded by the Korea government (MSIT) (No. RS-2025-16068196).

\bibliographystyle{splncs04}
\bibliography{main}
\end{document}


\title{MASS: Motion-Aligned Selective Scan for Refinement in Flow-Based Video Frame Interpolation \\
\textit{-Supplementary Material-}}
\titlerunning{MASS: Motion-Aligned Selective Scan for Refinement in VFI}

\author{Jun-Sang Yoo\orcidlink{0000-0001-9053-699X} \and
Seung-Won Jung\orcidlink{0000-0002-0319-4467}\thanks{Corresponding author.}}

\authorrunning{J.-S Yoo and S.-W. Jung}

\institute{Department of Electrical Engineering, Korea University, Seoul, Korea 
\email{\{junsang7777,swjung83\}@korea.ac.kr}}
\thispagestyle{empty}
\appendix
\maketitle

This supplementary is organized as follows.
{
\small
\begin{itemize}

    \item Sec.~\ref{sec:supp_mass_diagnostics} provides additional in-depth diagnostics of MASS to clarify where the gains arise across stages and regions.

    \item Sec.~\ref{pnp} evaluates the plug-and-play capability of MASS on external VFI backbones. 

    \item Sec.~\ref{xv} reports quantitative comparisons under an additional training-data protocol. 

    \item Sec.~\ref{eff} analyzes runtime and GPU memory usage.

    \item Sec.~\ref{ablation} studies the sensitivity to the adaptive trajectory budget and the velocity-aware step size. 

    \item Sec.~\ref{qual} presents more qualitative results including failure cases and additional comparisons.

    \item Sec.~\ref{detail} details the implementation, including the adaptive controller, variable-length trajectory buffering, refinement modules, and the synthesis network.
\end{itemize}
}

\begin{table*}[t]
    \centering
    \setlength{\tabcolsep}{2.0pt}
    \renewcommand{\arraystretch}{1.15}
    \caption{Stage-wise diagnostics of MASS on the SNU-FILM \textit{Hard} and \textit{Extreme} splits. We report the reductions in alignment and merged errors, PSNR gains, and adaptive controller statistics ($K$, $\Delta$) across different scales.}
    \label{tab:mass_diag_stage}
    \resizebox{\linewidth}{!}{%
    \begin{tabular}{llccccccc}
        \toprule
        \textbf{Stage} & \textbf{Subset} & \textbf{WarpErr} (F $\rightarrow$ $\hat{F}$) & \textbf{MergedErr} ($F,M$ $\rightarrow$ $\hat{F},\hat{M}$) & \textbf{$\Delta$PSNR$_{\text{merged}}$} & \textbf{$K$ (Mean)} & \textbf{$\Delta$ (Mean)} & \textbf{Corr($\|F\|, K$)} & \textbf{Corr($\|F\|$, $\Delta$)} \\
        \midrule
        1/8 (Coarse) & Hard & 0.0372 $\rightarrow$ 0.0211 (39.1\% $\downarrow$) & 0.0297 $\rightarrow$ 0.0174 (39.0\% $\downarrow$) & +4.19 dB & 4.47 & 0.620 & 0.620 & -0.653 \\
                     & Extreme & 0.0519 $\rightarrow$ 0.0363 (27.4\% $\downarrow$) & 0.0429 $\rightarrow$ 0.0320 (25.0\% $\downarrow$) & +2.11 dB & 4.38 & 0.611 & 0.691 & -0.715 \\
        \midrule
        1/4 (Fine)   & Hard & 0.0226 $\rightarrow$ 0.0204 (12.5\% $\downarrow$)& 0.0198 $\rightarrow$ 0.0171 (17.2\% $\downarrow$)& +1.35 dB & 4.57 & 0.547 & 0.826 & -0.878 \\
                     & Extreme & 0.0369 $\rightarrow$ 0.0356 (6.2\% $\downarrow$) & 0.0333 $\rightarrow$ 0.0315 (9.1\% $\downarrow$)& +0.51 dB & 4.53 & 0.554 & 0.816 & -0.891 \\
        \bottomrule
    \end{tabular}}
\end{table*}

\section{Additional In-Depth Analysis of MASS}
\label{sec:supp_mass_diagnostics}

\begin{table*}[t]
    \centering
    \setlength{\tabcolsep}{25.0pt}
    \renewcommand{\arraystretch}{1.15}
    \caption{Region-sliced diagnostics on difficult pixels. We evaluate the reductions in merged error specifically on the Top-20\% highest motion and Top-20\% highest uncertainty region across different scales.}
    \label{tab:mass_diag_region}
    \resizebox{\linewidth}{!}{%
\begin{tabular}{lccc}
\hline
\multirow{2}{*}{\textbf{Stage}} & \multirow{2}{*}{\textbf{Subset}} & \textbf{Top-20\% Motion}                                   & \textbf{Top-20\% Uncertainty}                     \\
                                &                                  & \textbf{MergedErr ($F,M$ $\rightarrow$ $\hat{F},\hat{M}$)} & \textbf{MergedErr ($F,M$ $\rightarrow$ $\hat{F},\hat{M}$)}    \\ \hline
\multirow{2}{*}{1/8 (Coarse)}   & Hard                             & 0.0414 $\rightarrow$ 0.0221 (41.9\% $\downarrow$)          & 0.0284 $\rightarrow$ 0.0164 (39.0\% $\downarrow$) \\
                                & Extreme                          & 0.0573$\rightarrow$ 0.0399 (26.7\% $\downarrow$)           & 0.0403 $\rightarrow$ 0.0295 (18.3\% $\downarrow$) \\ \hline
\multirow{2}{*}{1/4 (Fine)}     & Hard                             & 0.0232 $\rightarrow$ 0.0215 (7.5\% $\downarrow$)           & 0.0205 $\rightarrow$ 0.0173 (18.3\% $\downarrow$) \\
                                & Extreme                          & 0.0400 $\rightarrow$ 0.0389 (2.7\% $\downarrow$)           & 0.0322 $\rightarrow$ 0.0300 (11.3\% $\downarrow$) \\ \hline
\end{tabular}}
\end{table*}

We further analyze where MASS helps on SNU-FILM \textit{Hard} and \textit{Extreme}, which are dominated by large-motion cases. The main finding is that MASS plays different roles across scales: at the 1/8 stage, it mainly corrects gross motion misalignment, whereas at the 1/4 stage, it provides smaller but still consistent improvements around ambiguous blending regions such as occlusion boundaries.

For compact notation, let $F=(F_{t\to 0}, F_{t\to 1})$ and $M$ denote the intermediate bidirectional flow and blending mask before MASS-guided refinement, and let $\hat{F}=(\hat{F}_{t\to 0}, \hat{F}_{t\to 1})$ and $\hat{M}$ denote the corresponding refined outputs. We use three diagnostics. \textbf{WarpErr} measures the $\ell_1$ alignment error after backward warping $I_0$ to time $t$. \textbf{MergedErr} measures the $\ell_1$ error of the coarse blended frame before the final synthesis network. \textbf{$\Delta$PSNR$_{\text{merged}}$} measures the PSNR gain of the coarse blended frame after MASS-guided refinement. We additionally report the mean controller outputs $\bar{K}$ and $\bar{\Delta}$, together with their Pearson correlations with the flow magnitude $\|F\|$.

\begin{align}
\mathrm{WarpErr}(F)
&=
\mathbb{E}_{p}
\left[
\left\lVert
\overleftarrow{\mathcal{W}}(I_0, F_{t\to 0})(p) - I_t(p)
\right\rVert_{1}
\right], \\
I_{\text{base}}(F,M)
&=
M \odot \overleftarrow{\mathcal{W}}(I_0, F_{t\to 0})
+
(1-M) \odot \overleftarrow{\mathcal{W}}(I_1, F_{t\to 1}), \\
\mathrm{MergedErr}(F,M)
&=
\mathbb{E}_{p}
\left[
\left\lVert
I_{\text{base}}(F,M)(p) - I_t(p)
\right\rVert_{1}
\right], \\
\Delta \mathrm{PSNR}_{\text{merged}}
&=
\mathrm{PSNR}\!\left(I_{\text{base}}(\hat{F},\hat{M}), I_t\right)
-
\mathrm{PSNR}\!\left(I_{\text{base}}(F,M), I_t\right).
\end{align}

Table~\ref{tab:mass_diag_stage} shows that MASS is most effective at the coarse 1/8 stage, where it reduces both WarpErr and MergedErr and improves the coarse-frame PSNR. This indicates that MASS primarily corrects severe motion misalignment before the final synthesis stage. At the 1/4 stage, the remaining gains are smaller but still consistently positive, suggesting that MASS continues to refine already improved intermediate predictions rather than correcting gross motion errors again.

The controller statistics also support the intended adaptive behavior. Larger motion is associated with larger trajectory budgets and smaller step sizes, as reflected by the positive correlation between $\|F\|= \frac{1}2{}(\|F_{t\rightarrow0}\|_2 +\|F_{t\rightarrow1}\|_2)$ and $K$ and the negative correlation between $\|F\|$ and $\Delta$.

To localize these gains, we further evaluate two challenging pixel subsets. The \textbf{Top-20\% Motion} subset contains the 20\% of pixels with the largest flow magnitude $\|F\|$. The \textbf{Top-20\% Uncertainty} subset contains the 20\% of pixels with the largest mask ambiguity, where the uncertainty score is defined as
\begin{equation}
U(m)=4M(1-M).
\end{equation}

Table~\ref{tab:mass_diag_region} shows the distinct roles of each scale. At the coarse 1/8 stage, MASS produces its largest error reduction on the Top-20\% Motion regions, showing that it first corrects fast and severely misaligned motion. At the fine 1/4 stage, the relative gain becomes larger on the Top-20\% Uncertainty regions than on the Top-20\% Motion regions, indicating that the later stage is more useful for stabilizing ambiguous blending regions, such as occlusion boundaries.

\section{Plug-and-Play Capability of MASS}\label{pnp}
To assess the modular compatibility of MASS, we replace the intermediate motion aggregation/refinement branch in EMA-VFI-S and VFIMamba-S with MASS while keeping the original synthesis networks unchanged. We then retrain the modified models from scratch on Vimeo90K. The refined intermediate flows, occlusion masks, and motion-aligned aggregated features generated by MASS are directly fed into the original synthesis networks of the respective baselines.

As shown in Table \ref{tab:plug_and_play}, integrating MASS provides consistent improvements in high-resolution and large-displacement scenarios, while maintaining baseline performance on easier subsets. This suggests that the proposed module helps mitigate intermediate flow uncertainties and alignment errors, which frequently degrade reconstruction quality under complex dynamics. Overall, these results indicate that MASS can serve as a flexible plug-and-play component, allowing existing flow-based architectures to better handle challenging motions.

\begin{table}[t]
    \centering
    \renewcommand{\tabcolsep}{2mm}
\caption{Quantitative evaluation of MASS as a modular replacement for the intermediate motion aggregation/refinement branch in existing flow-based architectures (EMA-VFI-S and VFIMamba-S).}
    \label{tab:plug_and_play}
    \resizebox{\linewidth}{!}{
    
        \begin{tabular}{lcccccccc}
            \toprule
            \multirow{2}{*}{Model} & Training & 
            \multicolumn{4}{c}{SNU-FILM} &
            \multicolumn{2}{c}{Xiph} &
            \multirow{2}{*}{FLOPs (T)} \\
            \cmidrule(lr){3-6} \cmidrule(lr){7-8}
             & dataset & Easy & Medium & Hard & Extreme & 2K & 4K & \\
            \midrule
            EMA-VFI-S & V & 39.81/0.9903 & 35.88/0.9792 & 30.68/0.9371 & 25.47/0.8627 & 36.54/0.942 & 34.24/0.902 & 0.39 \\\rowcolor[gray]{.90}
            ~~~~ + MASS & V & 39.99/0.9907 & 35.92/0.9795 &30.79/0.9385 & 25.61/0.8639 & 36.63/0.942 & 34.36/0.903 & 0.57\\
            VFIMamba-S & V & 40.00/0.9907 & 35.81/0.9791 & 30.58/0.9378 & 25.30/0.8620 & 36.26/0.940 & 33.52/0.890 & 0.39\\\rowcolor[gray]{.90}
            ~~~~ + MASS & V & 39.96/0.9906 & 35.80/0.9791 & 30.69/0.9381 & 25.58/0.8635 & 36.54/0.942 & 34.04/0.899 & 0.54\\
            \bottomrule
        \end{tabular}
    }
\end{table}

\begin{table}[t]\caption{Quantitative comparison across benchmarks using the expanded V+X (Vimeo90K + X4K1000FPS) training dataset. The results for the baseline methods are taken directly from their original papers~\cite{liu2024sparse,zhang2024vfimamba}. (IE for Middlebury (M.B.); PSNR/SSIM for Vimeo90K, UCF101, Xiph, and SNU-FILM). } \label{tab:quantitativevx}
\centering
\resizebox{\linewidth}{!}{%
\begin{tabular}{lccccccccccccc}
\toprule
\multirow{2}{*}{Method}  & Training &\multirow{2}{*}{M.B.} & \multirow{2}{*}{Vimeo90K} & \multirow{2}{*}{UCF101} & \multicolumn{2}{c}{Xiph} & \multicolumn{4}{c}{SNU-FILM} & \multirow{2}{*}{Params (M)} \\ \cmidrule(lr){6-7} \cmidrule(lr){8-11}
  & Dataset& & & & 2K & 4K &Easy & Medium & Hard & Extreme    \\ \midrule

VFIMamba-S~\cite{zhang2024vfimamba} & V+X & 1.97 & 36.09/0.9800 & 35.35/0.9696 & 36.71/0.942 & 34.26/0.902 & 40.21/0.9912 & 36.17/0.9802 & 30.80/0.9382 & 25.59/0.8655 & 16.8
\\
SGM-VFI~\cite{liu2024sparse} & V+X & 1.87 & 35.81/0.9785 & 35.33/0.9692 & 36.06/0.940 & 33.26/0.897 & 40.36/0.9900 & 36.12/0.9787 & 30.62/0.9351 & 25.38/0.8615 & 20.8
\\

VFIMamba~\cite{zhang2024vfimamba} & V+X & 1.89 & 36.45/0.9807 & 35.37/0.9699 & 37.02/0.944 & 34.39/0.904 & 40.41/0.9903 & 36.30/0.9794 & 30.89/0.9387 & 25.68/0.8661 & 66.1
\\

\midrule


Ours-S & V+X & 1.90 & 36.30/0.9805 & 35.45/0.9700 & 36.90/0.944 & 34.72/0.908 & 40.25/0.9909 & 36.21/0.9803 & 30.96/0.9391 & 25.79/0.8671 & 16.5\\

Ours  & V+X & 1.78 & 36.54/0.9817 & 35.49/0.9703 & 37.11/0.947 & 34.87/0.908 & 40.36/0.9911 & 36.33/0.9808 & 31.16/0.9410 & 26.08/0.8663 & 67.4 \\
\bottomrule
\end{tabular}%
}
\end{table}

\section{Quantitative Comparison with Additional Training Data}\label{xv}
In the main paper, we constrained all evaluated models to be trained exclusively on the Vimeo90K~\cite{xue2019video} dataset to ensure a fair comparison. Since several recent methods~\cite{liu2024sparse, zhang2024vfimamba} were originally trained or fine-tuned with an additional dataset (X4K1000FPS~\cite{sim2021xvfi}), we re-trained them from scratch using only Vimeo90K for our primary evaluations.

In this section, to provide a direct and fair comparison against the official numbers reported in those recent papers, we adopt their expanded training protocol~\cite{liu2024sparse}. Specifically, we fine-tune our models using the additional dataset and denote this training setting as V+X in Table~\ref{tab:quantitativevx}.

As shown in the table, our models outperform recent state-of-the-art approaches under the same V+X training protocol. Notably, MASS exhibits distinct performance gains on datasets characterized by large motions and high resolutions. This indicates that our proposed architecture effectively leverages richer training data to further enhance its robustness in challenging scenarios.

\section{Efficiency and Memory Analysis} \label{eff}
We evaluate the inference runtime and GPU memory footprint of our models against recent VFI models on a single NVIDIA RTX 3090 GPU. To ensure accurate measurement, inference times are averaged over 100 iterations following 20 warm-up runs. All timings are recorded with \texttt{torch.cuda.synchronize()} to capture pure model execution time, excluding data loading overhead. 

As shown in Table~\ref{tab:efficiency_comparison}, Although Ours-S is not the lowest-latency model, it maintains a reasonable runtime/memory footprint while offering its clearest accuracy gains in the large-motion and high-resolution regime. 
Scaling to the full-capacity model (Ours) naturally increases the computational overhead due to the expanded channel dimensions. While the pixel-wise scanning mechanism introduces a slight runtime overhead relative to its low GPU memory footprint, the full model still demonstrates comparable overall efficiency.

To address hidden trajectory-construction overhead, we further detail the 720p module-wise runtime breakdown as follows: Extractor (14.9\%), Construction (9.5\%), VA-SSM (64.8\%), Refine (1.5\%), and Synthesis (9.3\%). Trajectory construction accounts for only 9.5\% of the total runtime, whereas the dominant cost comes from VA-SSM aggregation. These results show that iterative motion-aligned sampling is not a major computational bottleneck and clarify the primary computational cost of MASS.

\begin{table}[t]
    \centering
    \renewcommand{\tabcolsep}{4mm}
    \caption{
    Comparison of inference time (ms) and GPU memory usage (GB) across different resolutions. All measurements were conducted on a single NVIDIA RTX 3090 GPU. OOM is out of memory.}
    \label{tab:efficiency_comparison}
    \resizebox{\linewidth}{!}{

\begin{tabular}{lcccccc}
\hline
\multirow{2}{*}{Method} & \multicolumn{3}{c}{Runtime (ms)}                           & \multicolumn{3}{c}{GPU Memory (GB)}                                               \\ \cmidrule(lr){2-4} \cmidrule(lr){5-7}
                        & 640 $\times$ 480 & 1280 $\times$ 720 & 2048 $\times$ 1024 & 640 $\times$ 480 & 1280 $\times$ 720 & 2048 $\times$ 1024 \\ \hline
EMA-VFI-S               &   19               &42                   &96                    &     0.9             &    2.0               &      4.2                               \\
VFIMamba-S              &     40             &109                   &282                    &   1.0               &       2.2            &   4.2                      \\
SGM-VFI-1/2-S              &     67             &368                   &1661                    &    1.4              &      5.0             &       19.0                  \\
LC-Mamba-B              &    82              & 228                   &509                    &    2.3              &        6.1           &   14.8                                \\ \hline
Ours-S                  &                 66            &   175                &           424   &  1.2             &       3.0            &   5.6                               \\ \hline

EMA-VFI               &     47 & 126 & 288 & 1.6 & 3.7  & 7.6                          \\
VFIMamba              & 106 & 302 & 801 & 1.6 & 4.3 & 9.0                      \\
SGM-VFI-1/2              & 91   & 442 & -- & 1.6 & 5.3 & OOM                  \\
LC-Mamba-P              & 139   & 388 & 877 & 2.4 & 6.3 & 14.8                               \\ \hline
Ours                  & 128  & 376 & 871  & 2.0 & 4.8 & 10.6                                   \\ \hline
\end{tabular}
    }
\end{table}

\section{Ablation on Adaptive Trajectory Budget and Velocity-Aware Step Size}\label{ablation}

To validate our hyperparameter choices for the motion-adaptive discretization strategy, we analyze the impact of the adaptive trajectory budget ($K$) and the velocity-aware step size ($\Delta$). As shown in Table~\ref{tab:ablation_K_delta_unified}, limiting the maximum sampling budget leads to noticeable performance degradation in scenarios with large displacements and high resolutions, demonstrating that under-sampling trajectories in fast-moving regions impairs the model's ability to capture continuous dynamics. Conversely, expanding the upper bound yields negligible performance gains while noticeably increasing computational cost. Furthermore, raising the minimum budget results in a slight performance penalty despite the increased compute, indicating that forcing unnecessarily dense scans in simple or static regions is counterproductive. 

Regarding the velocity-aware step size, maintaining a relatively high minimum step size degrades performance in fast-motion scenarios, implying that large motions require finer state updates during the scan to properly preserve high-frequency details. On the other hand, lowering the minimum step size further does not provide additional gains and can slightly hurt some settings. Among the tested settings, the proposed ranges provide the best accuracy-efficiency trade-off.

\begin{table}[t]
    \centering
    \renewcommand{\tabcolsep}{3.5mm}
    \caption{{Ablation study for adaptive trajectory budget $K$ and velocity-aware $\Delta$.} 
    FLOPs are computed using the SNU-FILM-Hard (1280 $\times$ 720) dataset-average $K$.}
    \label{tab:ablation_K_delta_unified}
    \resizebox{\linewidth}{!}{
        \begin{tabular}{lcccccc}
            \toprule
            \multirow{2}{*}{Settings} & Training &
            \multicolumn{2}{c}{SNU-FILM} &
            \multicolumn{2}{c}{Xiph} &
            \multirow{2}{*}{FLOPs (T)} \\
            \cmidrule(lr){3-4} \cmidrule(lr){5-6}
            & dataset & Hard & Extreme & 2K & 4K & \\
            \midrule

            \multicolumn{7}{l}{\textbf{(A) $K$-range sweep (keep Vel-$\Delta$ fixed to [0.25,1.0])}} \\
            \midrule
            Adaptive $K$ [2,4]
            & V & 30.74/0.937 & 25.45/0.863  &36.60/0.942  & 34.36/0.903  & 0.49 \\

            Adaptive $K$ [2,12]
            & V & 30.89/0.938 & 25.73/0.866  & 36.84/0.944 & 34.65/0.907   & 0.55 \\

            Adaptive $K$ [4,8]
            & V & 30.87/0.938 & 25.70/0.866  & 36.84/0.944  & 34.62/0.906  & 0.52 \\

            Adaptive $K$ [2,8] (default)
            & V & \textbf{30.90}/\textbf{0.939} & \textbf{25.73}/\textbf{0.867} &
            \textbf{36.84}/\textbf{0.944} & \textbf{34.65}/\textbf{0.907} & 0.50 \\ 

            \midrule
            \multicolumn{7}{l}{\textbf{(B) $\Delta$-range sweep (keep adaptive $K$ fixed to [2,8])}} \\
            \midrule
            Vel-$\Delta$ [0.50,1.0]
            & V & 30.84/0.938 & 25.68/0.866 & 36.84/0.944  & 34.64/0.907  & 0.50 \\

            Vel-$\Delta$ [0.10,1.0]
            & V & 30.88/0.938 & 25.70/0.866 & 36.78/0.944 & 34.62/0.906  & 0.50 \\

            Vel-$\Delta$ [0.25,1.0] (default)
            & V & \textbf{30.90}/\textbf{0.939} & \textbf{25.73}/\textbf{0.867} &
            \textbf{36.84}/\textbf{0.944} & \textbf{34.65}/\textbf{0.907} & 0.50 \\

            \bottomrule
        \end{tabular}
    }
\end{table}

\section{More Qualitative Results}\label{qual}

\begin{figure}[t]
\centering
\includegraphics[width=\linewidth]{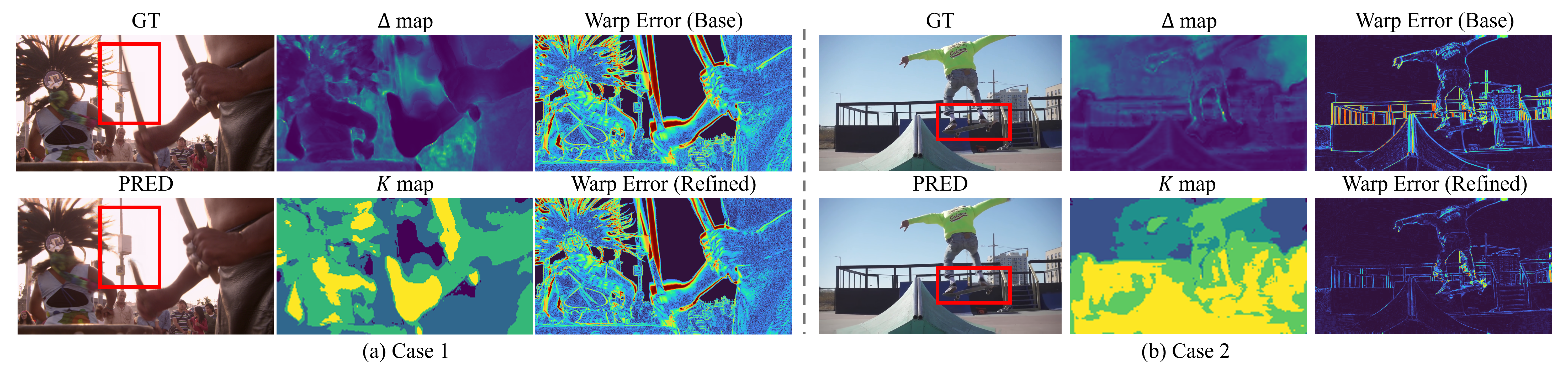}
\caption{Failure cases of MASS on challenging motion patterns. (a) In Case 1, despite meaningful responses from the adaptive control maps around the main subject, warp error reduction is limited, and the thin structure is partially lost. (b) In Case 2, while MASS improves the overall base warp error, scene-wide camera motion leads to a broad $K$-map activation, making the budget allocation less selective and leaving residual blur on fast-moving boundaries.}
\label{fig_sup_fail}
\end{figure}

\begin{figure}[t]
\centering
\includegraphics[width=\linewidth]{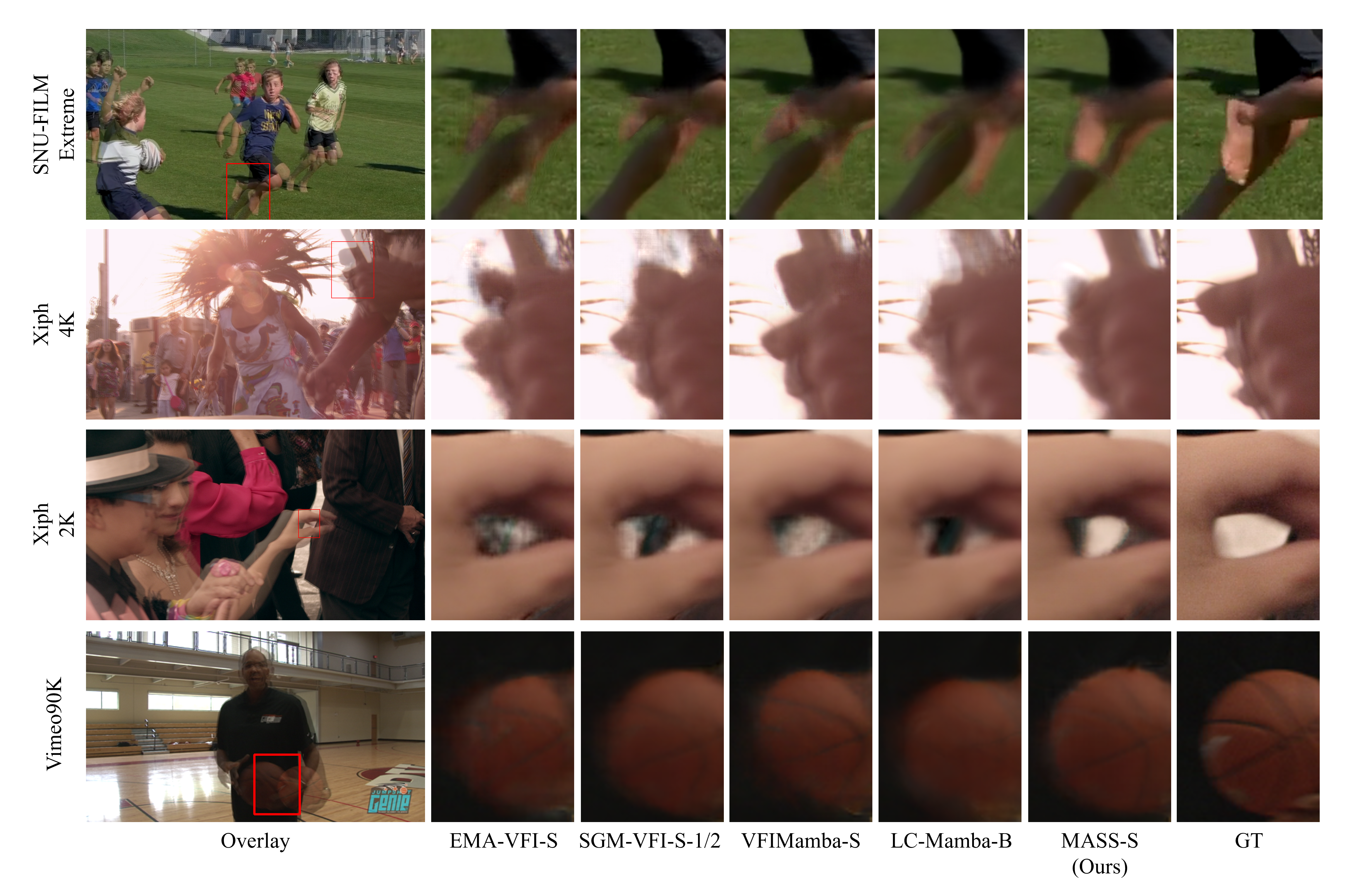}
\caption{Additional qualitative comparisons on the SNU-FILM, Xiph, and Vimeo90K datasets.}
\label{fig_sup_moreresults}
\end{figure}

Fig.~\ref{fig_sup_fail} shows two representative failure cases. In Case 1, the adaptive controller produces meaningful $K$ and $\Delta$ responses around the dominant moving body, indicating that the mechanism itself is active. Nevertheless, the reduction in warp error remains limited, and the thin structure inside the highlighted region is weakly reconstructed in the final prediction. This suggests a failure mode where extremely thin, short-lived structures remain difficult to recover even when motion-adaptive control is functioning.

In Case 2, MASS substantially reduces the base warp error across most of the frame, demonstrating effective correction for large portions of the scene. However, residual blur persists around the skateboard and lower-body boundaries. The broadly activated $K$-map indicates that under scene-wide motion—such as camera panning or strong parallax—the trajectory-budget allocation can become less spatially selective than ideal.

Fig.~\ref{fig_sup_moreresults} provides further visual comparisons of MASS-S against recent state-of-the-art methods across various resolutions. The results show that MASS reduces ghosting artifacts and preserves structural details compared to the other methods.

\section{Implementation Details}\label{detail}
We employ a Cosine Annealing scheduler, where the learning rate is linearly warmed up to $2 \times 10^{-4}$ for the first 2,000 steps and then gradually decayed to $1 \times 10^{-6}$ over 300 epochs.
During training, we apply comprehensive data augmentation, including random horizontal/vertical flips and temporal reversal. Input frames are randomly cropped to patches of size $256 \times 256$.
We instantiate two variants: the standard MASS ($C=32$) and the lightweight MASS-S ($C=16$), where $C$ denotes the base channel dimension of the initial feature extractor. All other architectural configurations are shared.

\subsection{Adaptive $K$ and velocity-aware $\Delta$}
In the main paper, we conceptually defined the dynamic sampling budget $K(p)$ using a scaling factor $\alpha$. In practice, to ensure numerical stability and reduce sensitivity to outlier flow magnitudes, we instantiate this scaling dynamically via mean-normalization and a rational squashing function.Given the average bidirectional flow magnitude $m(p) = \|F(p)\|$ (as defined in the main text), we first compute its frame-wise mean $\mu_m=\mathbb{E}_p[m(p)]$ (treated as a stop-gradient). We then normalize the magnitude:
\begin{equation}
    m_{\text{norm}}(p)=\frac{m(p)}{\mu_m+\epsilon}.
\end{equation}

The continuous sampling budget is then calculated by scaling within our predefined bounds:

\begin{equation}
    K_{\text{float}}(p)=K_{\min}+(K_{\max}-K_{\min})\frac{m_{\text{norm}}(p)}{m_{\text{norm}}(p)+1}.
\end{equation}

Finally, we obtain the integer budget via $K(p)=\text{clip}(\lfloor K_{\text{float}}(p) \rceil, K_{\min}, K_{\max})$. We set $[K_{\min}, K_{\max}] = [2, 8]$. This formulation preserves the proportional relationship intended by $\alpha$ while strictly bounding the behavior in edge cases.

Similarly, for the velocity-aware discretization introduced in the main paper, the normalization parameter $\beta$ is implemented adaptively based on the frame-wise mean velocity. Defining the trajectory's average step velocity as $v(p) = m(p) / K(p)$ and its spatial mean as $\mu_v=\mathbb{E}_{p}[v(p)]$ (stop-gradient), we compute a normalized velocity $v_{\text{norm}}(p) = v(p) / (\mu_v + \epsilon)$. The trajectory-specific step size is then:

\begin{equation}
\Delta(p)=\text{clip}\!\left(\frac{1}{1+v_{\text{norm}}(p)},\,\Delta_{\min},\Delta_{\max}\right),
\end{equation}
where $[\Delta_{\min}, \Delta_{\max}] = [0.25, 1.0]$. Since $\Delta(p)$ is a spatial map defined per-pixel, we broadcast it along the trajectory length and apply it as a uniform multiplier to the $dt$ pre-activation in the Mamba selective scan.

\subsection{Variable-length trajectories via fixed-length buffering and masking}
Each pixel corresponds to a trajectory sequence of variable length $K(p)+1$. To enable efficient batched computation, we allocate a fixed-length buffer of size $L=K_{\max}+1$ and define a validity mask:
\begin{equation}
    V(p,k)=\mathbb{I}[k\le K(p)],\quad k\in\{0,\dots,L-1\}.
\end{equation}

To avoid convolutional bleeding, we first zero out padded tokens using $V(p,k)$ before applying the depth-wise 1D convolution. Furthermore, we force the padded steps to approximate $\Delta\approx0$ by setting their $\Delta$ pre-activation to a large negative value (e.g., $-10^{4}$). Finally, we gather the output at the last valid index $k=K(p)$.

\begin{figure}[t]
\centering
\includegraphics[width=\linewidth]{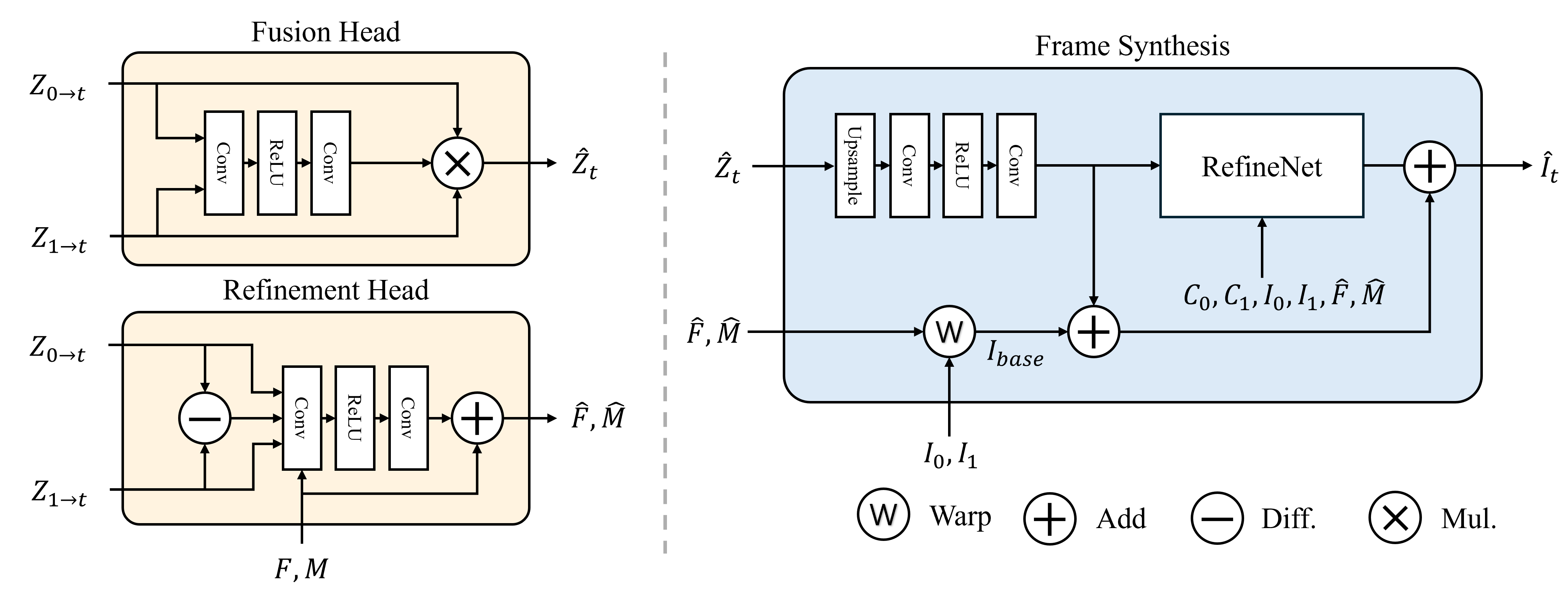}
\caption{Detailed architectures of the Fusion Head, Refinement Head, and Frame Synthesis network.}
\label{fig_sup_head}
\end{figure} 

\subsection{MASS-guided Refinement}
At each scale, we utilize two dedicated heads for context fusion and motion correction. The fusion head (Fig.~\ref{fig_sup_head}, top left) merges the bidirectional motion-aligned contexts ($Z_{0\to t}, Z_{1\to t}$) into a unified representation $\hat{Z}_t$ via a gating mechanism. Concurrently, a lightweight refinement head (Fig.~\ref{fig_sup_head}, bottom left) corrects the intermediate flow and mask. It consumes the bidirectional contexts, their absolute discrepancy ($|Z_{0\to t}-Z_{1\to t}|$), and the current estimates ($F, M$) to predict residual updates ($\Delta F, \Delta M$) and a sigmoid gate $u \in (0,1)$. The refined estimates are updated via residual addition:
\begin{equation}
    \hat{F} = F + u \cdot \Delta F, \quad \hat{M} = M + u \cdot \Delta M.
\end{equation}

The final convolutional layers of the Refinement Head are zero-initialized to ensure the update initially acts as an identity mapping.

\subsection{Frame synthesis network $G$}
As illustrated in Fig.~\ref{fig_sup_head} (right), we adopt a coarse-to-fine synthesis strategy. We first warp and blend the input frames ($I_0, I_1$) using the refined flow and mask ($\hat{F}, \hat{M}$) to form a structural base image $I_{base}$. Concurrently, the fused context $\hat{Z}_t$ from the fusion head is spatially upsampled, processed via shallow convolutions, and added to $I_{base}$ to inject high-frequency details. Finally, a RefineNet~\cite{huang2022real} takes this intermediate representation alongside the source inputs, local features ($C_0, C_1$), and refined motions to predict the final RGB residual, yielding the interpolated frame $\hat{I}_t$.

\bibliographystyle{plain}
\bibliography{main}